\documentclass[10pt,twocolumn,letterpaper]{article}

\usepackage{cvpr}
\usepackage{times}
\usepackage{epsfig}
\usepackage{graphicx}
\usepackage{amsmath}
\usepackage{amssymb}

\usepackage{times}
\usepackage{epsfig}
\usepackage{graphicx}
\usepackage{amsmath}
\usepackage{amssymb}
\usepackage{color}
\usepackage{algorithm, algpseudocode}
\usepackage{amsmath,amssymb}
\usepackage{sidecap}
\usepackage{caption}
\usepackage{enumitem}
\usepackage{subfigure}
\usepackage{bm}
\usepackage{colortbl}
\usepackage{hhline}

\definecolor{color3}{rgb}{0.95,0.95,0.95}
\definecolor{color4}{rgb}{0.96,0.96,0.86}
\definecolor{color1}{rgb}{0.90,0.94,0.84}
\definecolor{color2}{rgb}{1,0.92,0.8}

\newcommand{\best}[1]{\cellcolor{color1}{{#1}}}%
\newcommand{\sbest}[1]{\cellcolor{color2}{{#1}}}%

\usepackage{tabularx}
\newcolumntype{b}{>{\centering\arraybackslash}X}
\newcolumntype{d}{>{\raggedleft\arraybackslash}X}
\newcolumntype{e}{>{\raggedright\arraybackslash}X}
\newcolumntype{s}{>{\hsize=.5\hsize}X}
\newcolumntype{m}{>{\hsize=.35\hsize}b}
\newcolumntype{z}{>{\hsize=.75\hsize}b}

\usepackage{multirow}
\usepackage{nicefrac}     
\usepackage{fixltx2e}
\usepackage{array}

\let\oldReturn\Return
\renewcommand{\Return}{\State\oldReturn}

\usepackage[pagebackref=true,breaklinks=true,letterpaper=true,colorlinks,bookmarks=false]{hyperref}

\cvprfinalcopy 

\begin{document}

\title{Task-generalizable Adversarial Attack based on Perceptual Metric}

\author{Muzammal Naseer$^{\dagger}$, Salman H. Khan$^{\star}$$^{\ddagger}$, Shafin Rahman$^{\dagger}$$^{\star}$, Fatih Porikli$^{\dagger}$   \\
 $^{\dagger}$Australian National University, $^{\star}$Data61-CSIRO, $^{\ddagger}$Inception Institute of AI\\
{\tt\small muzammal.naseer@anu.edu.au}
}

\maketitle


\begin{abstract}
Deep neural networks (DNNs) can be easily fooled by adding human imperceptible perturbations to the images. These perturbed images are known as `adversarial examples' and pose a serious threat to security and safety critical systems. A litmus test for the strength of adversarial examples is their transferability across different DNN models in a black box setting (\ie when the target model's architecture and parameters are not known to attacker). 

Current attack algorithms that seek to enhance adversarial transferability work on the decision level \ie generate perturbations that alter the network decisions. This leads to two key limitations: (a) An attack is dependent on the task-specific loss function (\eg softmax cross-entropy for object recognition) and therefore does not generalize beyond its original task. (b) The adversarial examples are specific to the network architecture and demonstrate poor transferability to other network architectures. 

We propose a novel approach to create adversarial examples that can broadly fool different networks on multiple tasks. Our approach is based on the following intuition: ``Perpetual metrics based on neural network features are highly generalizable and show excellent performance in measuring and stabilizing input distortions. Therefore an ideal attack that creates maximum distortions in the network feature space should realize highly transferable examples". We report extensive experiments to show how adversarial examples generalize across multiple networks for classification, object detection and segmentation tasks.
\end{abstract}

\begin{figure}[!t]
  \centering
    \includegraphics[trim= 8mm 0mm 14mm 0mm, clip, width=1\linewidth]{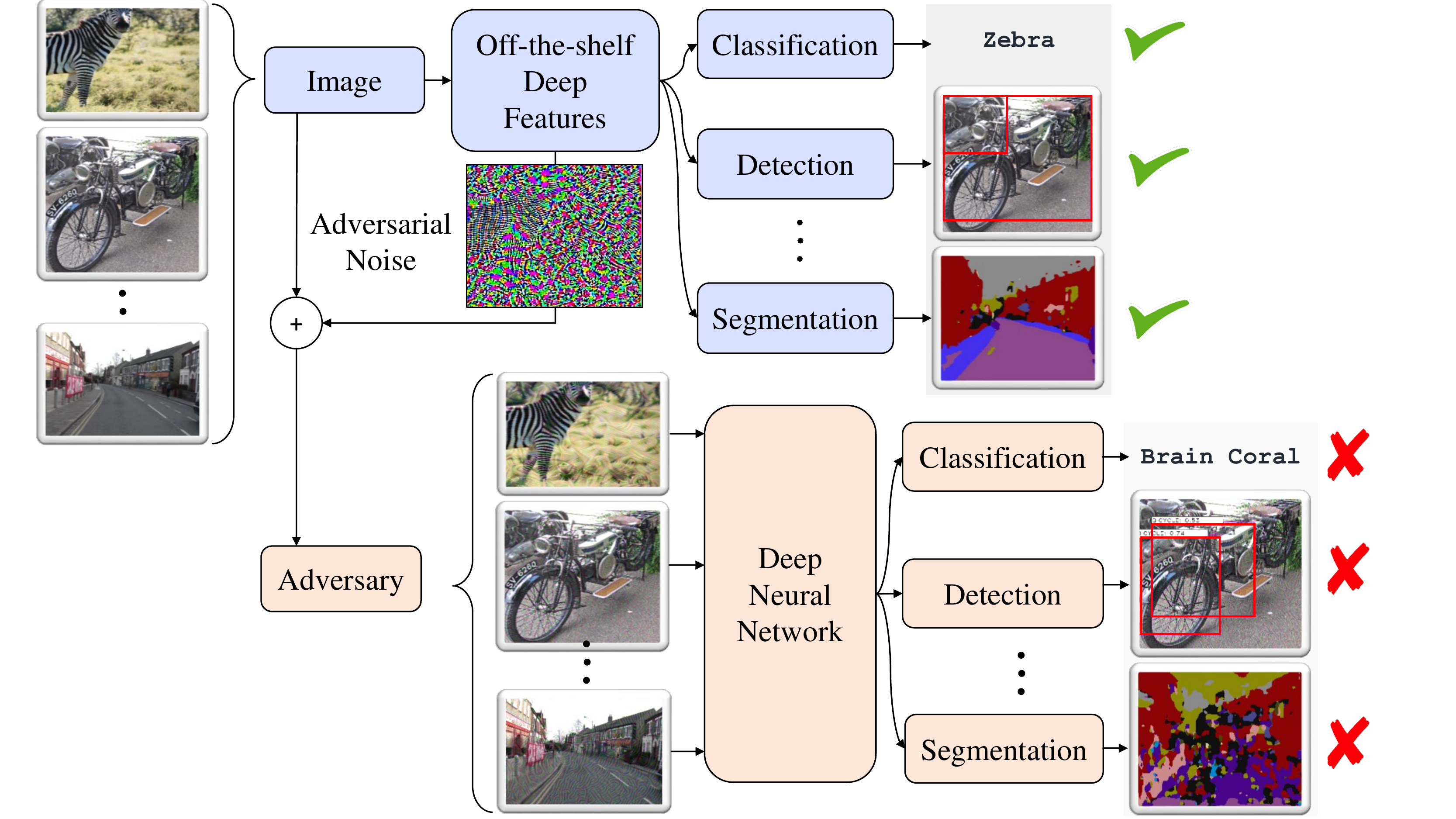}
    \caption{Similar to off-the-shelf deep features that are employed to boost the performance of different computer vision tasks, adversarial noise patterns found in deep features space are transferable across different tasks. (\emph{Noise pattern is magnified for better visualization})
    }
    \label{fig:concept_diagram}
    \vspace{-2ex}
\end{figure}

\section{Introduction}
Transferability is a phenomenon where adversarial examples created for one network can fool others. The transferability of adversarial examples makes it challenging to deploy deep neural networks in security critical environments.  This is of high concern because it gives attackers the flexibility to train a local network and transfer its attack against an already deployed network, without knowing its architecture or parameters (`\emph{black-box attacks}'). Current attack algorithms \cite{Kurakin2016AdversarialEI, carlini2017towards} perform well when the network architecture and parameters are known (`\emph{white-box setting}'); however, their strength significantly decreases in the black box setting, as shown in \cite{Su2018IsRT}. Recent attempts on enhancing the transferability in black-box settings have been reported in \cite{Dong_2018_CVPR,xie2018improving, Zhou_2018_ECCV}. Nevertheless, their dependency on a task-specific loss function make them non-transferable across different tasks. For example, to fool classification models, the attacker starts from the softmax cross-entropy to find a gradient direction that increases the model loss for a given sample. Examples found in this way are specific and do not generalize beyond their original task.

We propose a novel approach to generate high strength adversarial examples that are transferable across different network architectures and, most importantly, across different vision tasks (\eg, image segmentation, classification and object detection). Our approach is based on the following intuitions: \textbf{(a)} neural networks trained on ImageNet \cite{ILSVRC15} (or other sufficiently large image datasets) learn generic internal representations that are transferable to new tasks and datasets \cite{kornblith2018better,sharif2014cnn}.  As a result, it is common practice to use pre-trained classification networks as the basic building block (\emph{network backbone}) for a variety of different tasks \cite{lin2018focal,long2015fully} and, \textbf{(b)} a perceptual metric based on VGG internal representations aligns well with human perception \cite{Zhang_2018_CVPR} and can be used not only to measure the input distortion but also to stabilize it \cite{Kupyn_2018_CVPR}. We hypothesize that adversarial examples based on perceptual distortion, under a given bound, \eg $l_{\infty} \le \epsilon$, in the deep features space, are ideally suited to fool any deep network, whether designed for classification, object detection, segmentation or other vision tasks. We present the first such algorithm, which creates adversarial examples by distorting the deep neural activations. This not only generates high-strength perturbations but also provides flexibility to work with any task, as the proposed attack does not use any task-dependent loss function.

To the best of our knowledge, the closest to our approach is a decision-boundary free attack (called FFF) \cite{mopuri-bmvc-2017}. The idea is to train a single perturbation within a given metric norm to maximize the activation response of the network's internal layers. After training, the perturbation is added to the input images to make them adversarial. The problem with this approach is that it optimizes adversarial noise in a way that is independent of the data sample; hence noise, severely overfits the network and has very low transferability. In contrast, we do not optimize a single noise pattern, instead we directly maximize the distortions in the network's internal representations for a given input sample. Zhou \etal \cite{Zhou_2018_ECCV} also proposed to maximize representation loss. However, their approach is specific to only the classification task, since the gradient direction in their approach is dependent on cross-entropy loss, which requires labels for the task at hand. Thus, their attack algorithm is essentially a \emph{supervised} adversarial attack. In contrast, we don't use any task dependent loss in our objective, so our attack method does not rely on any labels. Thus, it is an \emph{unsupervised} adversarial attack. Furthermore, we focus on VGG networks for high-dimensional datasets. Remarkably, other networks (Inception/Resnet) do not offer enough distortion in a constrained optimization scenario to carry out this type of attack.

One intriguing aspect of our approach is its simplicity and efficiency. For instance, we only use features from a single layer (conv3.3) of VGG-16 \cite{Simonyan14verydeep} (instead of multiple layers as in \cite{mopuri-bmvc-2017}) and calculate the mean squared difference between the original and adversarial examples to represent neural representation distortion (NRD). NRD is fully differentiable and its minimization can help in image restoration problems \cite{Kupyn_2018_CVPR}. Here, we propose to maximize the NRD to construct   adversarial examples. Finding adversarial examples based on feature representation makes our attack generalizable across different architectures for different tasks. Specifically, we show high inter-task and intra-task transferability for our approach on large-scale datasets, including ImageNet \cite{ILSVRC15}, MS-COCO \cite{lin2014microsoft} and CAMVID \cite{brostow2009semantic}. 

Our method is not restricted to the original backbone models trained on a specific benchmark. Most backbone models are fine-tuned with additional training datasets to a specific task. As we elaborate in Sec.~\ref{sec: results}, our method can successfully be applied to any network that is pretrained on one benchmark, then fine-tuned on another, e.g. RetinaNet \cite{lin2018focal} and SegNet \cite{Badrinarayanan2017SegNetAD}.

\textbf{Contributions:} We study and highlight the importance of a neural network's internal representations (Fig. \ref{fig:concept_diagram}) in the context of adversarial attacks. Our major contributions are:
\begin{itemize}[noitemsep,topsep=0pt]
\item We propose a generalizable, black-box, untargeted adversarial attack algorithm on a neural network's internal representation. 
\item We leverage generic representations learned by models (e.g. VGG-16~\cite{Simonyan14verydeep}) trained on large image datasets (e.g. ImageNet ~\cite{ILSVRC15}) to construct transferable adversarial examples.
\item Our attack algorithm does not rely on a task-specific loss function or a specific set of input labels, therefore it demonstrates cross-network, cross-dataset, and cross-task transferability. 
\item We provide state-of-the-art results for classification networks and provide a robust benchmark to measure the robustness of any neural network based vision system against generic adversarial examples.
\end{itemize}

\section{Related Work}

Since the seminal work of Szegedy et al. \cite{szegedy2013intriguing} many adversarial attack algorithms \cite{goodfellow2014explaining, Kurakin2016AdversarialEI, athalye2017synthesizing, Dong_2018_CVPR} have been proposed to show the vulnerability of neural networks against imperceptible changes to inputs. A single-step attack, called fast gradient sign method (FGSM), was proposed by \cite{goodfellow2014explaining}. In a follow-up work, Kurakin et al. \cite{Kurakin2016AdversarialEI} proposed a robust multi-step attack, called iterative fast gradient sign methods (I-FGSM) that iteratively searches the loss surface of a network under a given metric norm. 
To improve transferability, a variant of I-FGSM, called momentum iterative fast gradient sign method (MI-FGSM), was introduced \cite{Dong_2018_CVPR}, which significantly enhances the transferability of untargeted attacks on ImageNet \cite{ILSVRC15} under a perturbation budget of $l_\infty \le 16$. Authors \cite{Dong_2018_CVPR} associated the transferability of MI-FGSM with its ability to break local maxima as the number of attack iterations increase. Recently, \cite{xie2018improving} proposed a data augmentation technique to further boost the transferability of these attack methods. In contrast to ours, all of these methods are supervised adversarial attacks dependent on cross-entropy loss to find the harmful gradient direction.

\begin{figure}[!t]
  \centering
    \includegraphics[trim= 7mm 2mm 14mm 12mm, clip, width=.80\linewidth]{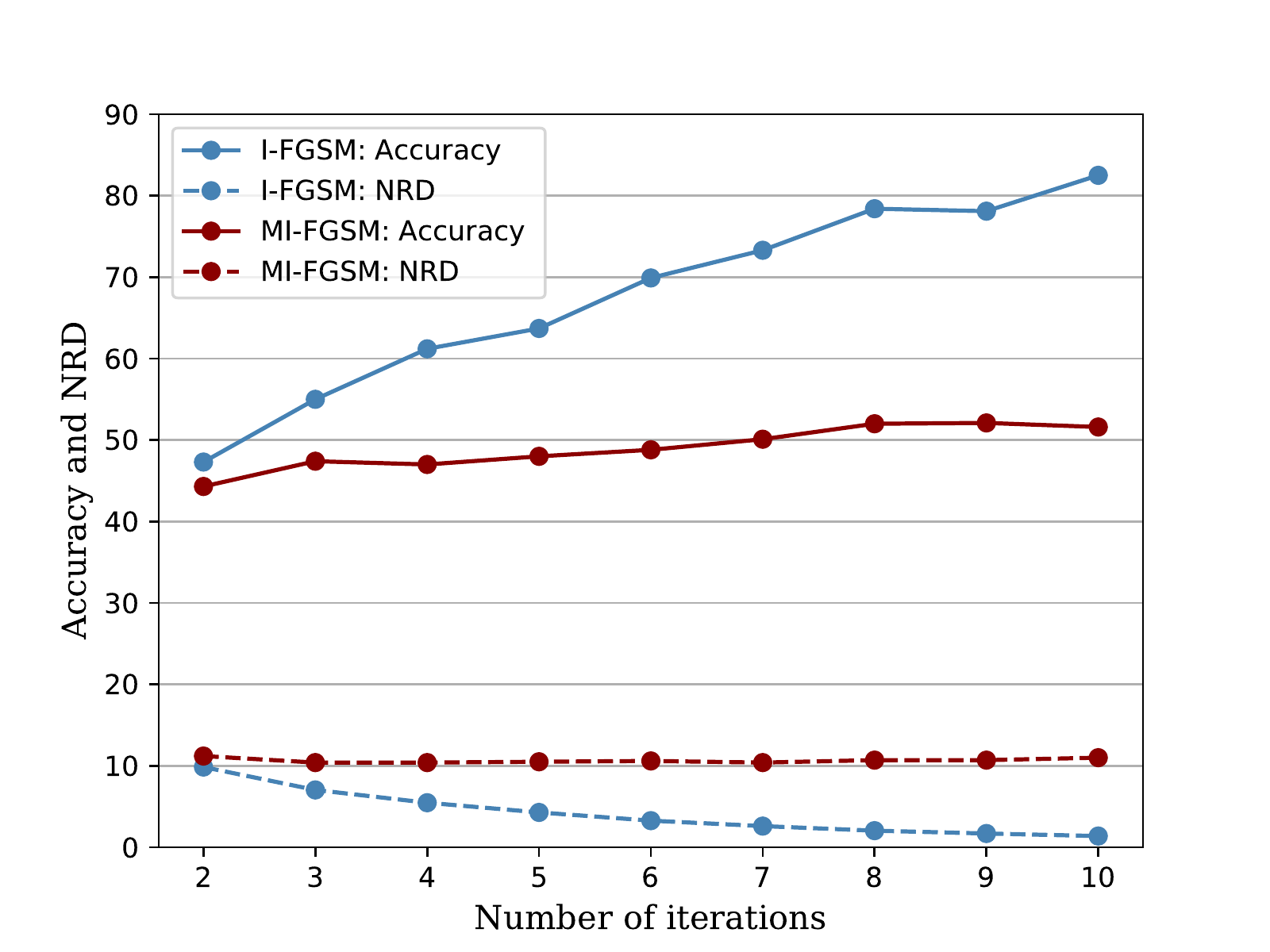}
    \caption{Accuracy of Inc-v4 and NRD is shown for adversarial examples generated on Inc-v3 by I-FGSM and MI-FGSM. NRD is averaged over all examples. As the number of iterations increases, the accuracy of Inc-v4 on adversarial example found by I-FGSM increases, i.e., the transferability of I-FGSM decreases along with its NRD.}
    \label{fig:iterations}
    \vspace{-2ex}
\end{figure}

Interestingly, NRD of I-FGSM decreases as the number of attack iterations increases as compared to MI-FGSM as shown in Fig.~\ref{fig:iterations}. We generate adversarial examples on ImageNet \cite{ILSVRC15} subset provided by the NIPS security challenge 2017. As can be seen, MI-FGSM maintains its NRD with increasing number of iterations. This also indicates that directly maximizing the NRD can boost the transferability of adversarial examples.

\begin{figure*}
\centering
  \begin{minipage}{.19\textwidth}
  	\centering
    Original
    \includegraphics[width=\linewidth, keepaspectratio]{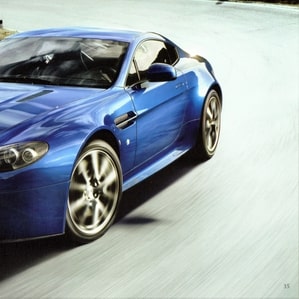}\
     (a) \small Sports Car
  \end{minipage}
   \begin{minipage}{.19\textwidth}
   	\centering
    $l_\infty \le 16$
    \includegraphics[width=\linewidth,keepaspectratio]{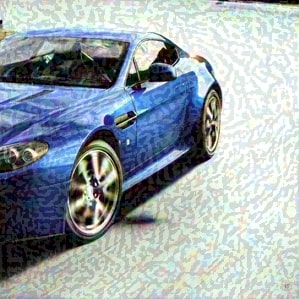}\
      (b) \small Racer
  \end{minipage}
  \begin{minipage}{.19\textwidth}
  	\centering
    $l_\infty \le 16$
    \includegraphics[width=\linewidth, keepaspectratio]{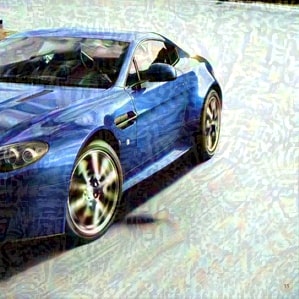}\
     (c) \small Racer
  \end{minipage}
  \begin{minipage}{.19\textwidth}
  	\centering
    $l_\infty \le 16$
     \includegraphics[width=\linewidth, keepaspectratio]{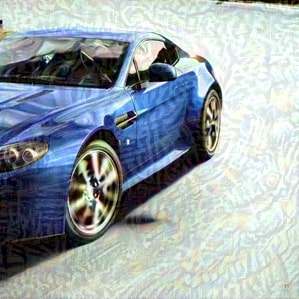}\
    (d) \small Loggerhead
  \end{minipage}
    \begin{minipage}{.19\textwidth}
  	\centering
    $l_\infty \le 16$
     \includegraphics[width=\linewidth, keepaspectratio]{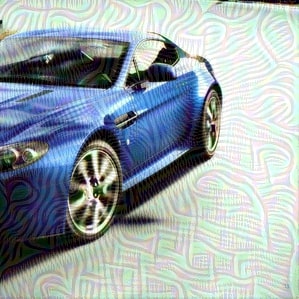}\
    (e) \small Quilt
  \end{minipage}
  \caption{VGG-16 output is shown for sample images. (a) represents benign example, while (b), (c), (d) and (e) show adversarial examples generated by FGSM, MI-FGSM, DIM and NRDM, respectively, against VGG-16 . All adversarial examples have distance $l_\infty \le 16$ from the original seed (a).}
\label{fig:L_inf_images}
\end{figure*}

\section{Adversarial Attacks}
\label{sec:attacks}
In this section, we first provide our problem setting, followed by a brief background to adversarial attacks. We explain how popular attack mechanisms, such as FGSM \cite{goodfellow2014explaining}, I-FGSM \cite{Kurakin2016AdversarialEI} and MI-FGSM \cite{Dong_2018_CVPR}, differ from each other. This background will form the basis of our proposed attack in Sec.~\ref{sec:NRD}.

\textbf{Problem Setting:}
In this paper, we specifically consider the transferability of untargeted attacks under the $l_\infty$ norm constraint on perturbation strength. Untargeted attacks are considered because they have higher transferability compared to targeted attacks \cite{Dong_2018_CVPR, xie2018improving}. Furthermore, to make sure that the benign and adversarial examples are close to each other, an attacker is constrained under a metric norm  like $l_\infty \le \epsilon$, i.e., in the case of images the attacker can change each pixel intensity value by at maximum $\epsilon$ amount. 

\subsection{FGSM} 
Adversarial examples can be formulated as a constrained optimization problem.  Suppose we are given a classifier function $\mathcal{F}$ that maps an input $\bm{x}$ to its ground-truth class $y$, a cost function $J(\bm{x},y)$  that is used to train the classifier and an allowed perturbation budget `$\epsilon$'.  FGSM \cite{goodfellow2014explaining} finds an adversarial example $\bm{x'}$ that satisfies $\parallel \bm{x'} - \bm{x} \parallel_{\infty} \le \epsilon$ using the following formulation:
\begin{equation}
\bm{x'} = \bm{x} + \epsilon\cdot\mathrm{sign}(\nabla_{\bm{x}}J(\bm{x},y)),
\label{eq:fgsm}
\end{equation}
where $\nabla_{\bm{x}}J(\bm{x},y)$ represent the gradient of the cost function w.r.t input $\bm{x}$. A common choice for $J$ is the cross-entropy loss.
The problem with FGSM is that it is a single-step attack, which reduces the attack success rate due to underfitting the threat model. To overcome this difficulty, an iterative version of FGSM was proposed \cite{Kurakin2016AdversarialEI}.

\subsection{I-FGSM}
I-FGSM \cite{Kurakin2016AdversarialEI} iteratively applies FGSM with a small step size $\alpha$ for a given number of iterations $T$. The step size $\alpha$ can be calculated by dividing the perturbation budget $\epsilon$ with the number of iterations $T$, i.e., $\alpha = \epsilon/T$. I-FGSM can be represented as follows for steps $t\in [1,T]$:
\begin{equation}
\bm{x'}_0 = \bm{x},\;\; \bm{x'}_{t+1} = \bm{x'}_t + \alpha\cdot\mathrm{sign}(\nabla_{\bm{x}}J(\bm{x'}_t,y)). 
\label{eq:ifgsm}
\end{equation}
\noindent The problem with I-FGSM is that it overfits the threat model, reducing model accuracy to even 0\%, while producing a small neural representation distortion (NRD) (See Fig.~\ref{fig:iterations} for empirical evidence). One side effect of having low NRD is the reduced transferability of adversarial examples. This is what Dong \etal \cite{Dong_2018_CVPR} built upon, proposing an attack algorithm that finds adversarial examples iteratively, while maintaining the transferability rate.

\subsection{MI-FGSM}
The work in \cite{Dong_2018_CVPR} added momentum into the optimization objective of I-FGSM. It can be expressed as follows:
\begin{align}
\label{eq:mifgsm}
\bm{x'}_0 =& \bm{x}, \;\; \bm{x'}_{t+1} = \bm{x'}_t + \alpha\cdot\mathrm{sign}(\bm{g}_{t+1}), \; t\in[1,T] \notag \\
&\bm{g}_{t+1} =  \mu \cdot \bm{g}_{t} + \frac{\nabla_{\bm{x}}J(\bm{x'}_{t},y)}{\|\nabla_{\bm{x}}J(\bm{x'}_{t},y)\|_1}.
\end{align}
The strength of MI-FGSM can be described by two of its control parameters, number of iterations and momentum. The number of attack iterations makes it strong in white-box settings (like I-FGSM), while momentum allows it to maintain NRD, enhancing the attack success rate in black-box settings.

Based on the above observations, we build our framework and propose to enhance the NRD directly to create strong adversarial examples for black-box attacks.

\section{Neural Representation Distortion}
\label{sec:NRD}

\textbf{The Problem:} Strong white-box attack algorithms \cite{Kurakin2016AdversarialEI, carlini2017towards} consider already-known network parameters $\bm{\theta}$ and perturb the input to create $\bm{x'}$, such that the example is misclassified, \ie, $\mathcal{F}(\bm{x'}; \bm{\theta})\neq y$. Since the perturbations are calculated using gradient directions that are specific to $\bm{\theta}$, the resulting perturbed images $\bm{x'}$ do not generalize well to other networks \cite{Dong_2018_CVPR, Su2018IsRT}. 
The attacks presented in \cite{Dong_2018_CVPR,xie2018improving, Zhou_2018_ECCV} show relatively better transferability, however, these attacks also perturb input images along gradient directions $\nabla_{\bm{x}}J$ that are dependent on the ground-truth label $y$ and the definition of the loss function $J$. This dependency limits the cross-network and cross-task transferability of these attacks. 

\textbf{Our Solution:} In this paper, we propose to directly maximize the perceptual metric based on representation loss of deep feature activations by solving the following optimization problem:
\begin{align}\label{eq: nrd}
 \underset{\mathbf{\mathbf{x}'}}{\text{max}} & \;\; \mathcal{F}(\bm{x}')|_{k} - \mathcal{F}(\bm{x})|_{k} \notag \\
 \text{subject to:} & \;\;  \Vert \bm{x} - \bm{x}' \Vert_{\infty} \leq \epsilon,
\end{align}
where $\mathcal{F}$ is DNN based classifier, $k$ is the internal representation layer and $\epsilon$ is the allowed perturbation budget. We apply a transformation $\mathcal{T}$ to input $\bm{x}$ at the first iteration (Algorithm \ref{alg:NRD}) to create a neural representation difference of an adversarial w.r.t a benign example and then maximize the mean squared error of this difference with in a given perturbation budget. There can be different choices for $\mathcal{T}$ but in this work $\mathcal{T}$ simply adds random noise to the input sample, i.e our algorithm takes a random step at the first iteration. Random noise is convenient to attain a difference at the starting point of our algorithm and it is preferable to heuristic transformations that may cause methodical bias.  

We use the VGG-16 \cite{Simonyan14verydeep} conv3.3 feature map as the neural representation distortion. This choice is based on observations, reported in the recent study \cite{Su2018IsRT}, that adversarial examples found in VGG space have high transferability. This is also evident in our experimentation (Table \ref{tab:untarget_imagenet}). Increasing the representation loss at multiple network layers did not notably increase attack success and adds a significant computational overhead. Our attack algorithm does not rely on the cross-entropy loss or input labels. This makes it a generic algorithm, which can be used to attack any system using off-the-shelf features in their pipeline. This makes several popular computer vision tasks vulnerable to adversarial attacks, e.g., object detection and segmentation. Furthermore, our proposed approach is complementary to recent best-performing attack methods, such as MI-FGSM \cite{Dong_2018_CVPR} and DIM \cite{xie2018improving}. Therefore, we demonstrates that it can be used alongside them, which further boosts the strength of adversaries. Our proposed method to maximize NRD for a given input sample is summarized in Algorithm \ref{alg:NRD}.
\begin{algorithm}[h]
\small
\caption{Neural Representation Distortion Method}
\label{alg:NRD}
\begin{algorithmic}[1]
\Require A classifier $\mathcal{F}$, input sample $\bm{x}$, input transformation $\mathcal{T}$, internal network layer $k$, perturbation budget $\epsilon$ and number of iterations $T$.
\Ensure
An adversarial example $\bm{x'}$ with $\|\bm{x'} - \bm{x}\|_{\infty} \leq \epsilon$.
\State $\bm{g}_0 = 0$; $\bm{x'} = \bm{x}$;
\For {$t = 0$ to $T-1$}
\If{$t=0$}
\State $\bm{x'}$ = $\mathcal{T}(x)$ 
\EndIf
\State Forward pass $\bm{x'}_t$ to $\mathcal{F}$ and compute $\mathcal{L}$ as follows;
\begin{equation}
\mathcal{L} = \|\mathcal{F}(\bm{x'})|_{k} - \mathcal{F}(\bm{x})|_{k}\|_{2};
\end{equation}
\State Compute gradients $\bm{g}_{t} = \nabla_{\bm{x}}\mathcal{L}(\bm{x'}_t, \bm{x})$;
\State Apply the following equation;
\vspace{-2ex}
\begin{equation}
\bm{x'}_{t+1} = \bm{x'}_{t} +\epsilon\cdot\mathrm{sign}(\bm{g}_{t});
\end{equation}
\vspace{-4ex}
\State Project adversary into the vicinity of $\bm{x}$
\vspace{-2ex}
\begin{equation}
\bm{x'}_{t+1} = clip(\bm{x'}_{t+1}, \hspace{1ex} \bm{x}-\epsilon,  \hspace{1ex} \bm{x}+\epsilon);
\end{equation}
\vspace{-4ex}
\EndFor
\Return $\bm{x'} = \bm{x'}_T$.
\end{algorithmic}
\end{algorithm}

\section{Experiments}
\label{sec:Expts}

\subsection{Evaluation Protocol}
In this section, we describe the datasets used for evaluation, network architectures under attack, and the parameter settings for each attack algorithm.

\subsubsection{Datasets}
We use the MNIST, and CIFAR10 test sets and the ImageNet \cite{ILSVRC15} subset provided by NIPS security challenge 2017 (ImageNet-NIPS) to validate the effectiveness of the proposed attack against classification models. The MNIST and CIFAR10 test sets contain 10k samples each, while ImageNet-NIPS contains 1k image samples. For object detection, we used MS-COCO \cite{lin2014microsoft} validation set, which contains 40.5k images. This is a multi-task dataset popular for image segmentation, object detection and image captioning tasks. We report adversarial attack performance against object detection, however adversarial examples found on this dataset can be used to fool other related tasks e.g., visual question answering. For segmentation, we use the CAMVID \cite{brostow2009semantic} test set to measure segmentation robustness against NRDM (Algorithm \ref{alg:NRD}). This dataset contains 233 image samples extracted from video sequences of driving scenes. 

\begin{table}[!h]
\small
\begin{center}
\begin{tabular}{|p{16ex}<{\raggedright}|p{22ex}<{\raggedright}|}
\hline
\rowcolor{color3}
\texttt{model-m} & \texttt{model-c}\\
\hline
conv2d(32, 3x3) & 2$*$\{conv2d(96, 3x3)\} \\
maxpool(2x2) & conv2d(96, 3x3, s=2)\\
conv2d(64, 3x3) & 2$*$\{conv2d(192, 3x3)\}\\
maxpool(2x2) &conv2d(96, 3x3, s=2) \\
\textbf{conv2d(64, 3x3)} & 2$*$\{\textbf{conv2d(192, 3x3)}\} \\
fc(64) & conv2d(10, 3x3) \\
softmax(10) & avg-pool \\
 & softmax(10) \\
\hline
\end{tabular}
\end{center}
\vspace{-3ex}
\caption{Architectures of naturally trained convolutional networks for MNIST (\texttt{model-m}) and CIFAR10 (\texttt{model-c}). `*' indicates the number of times a layer is repeated. `s' represent stride. Each convolutional layer is followed by ReLU activation. Batch-norm is used after each convolutional layer in \texttt{model-c}. Layers whose outputs are used by NRDM are highlighted in bold.}
\label{tab:mnist_cifar_conv_models}
\end{table}

\begin{table}[t]
\small
\begin{center}
\begin{tabular}{|p{26ex}<{\raggedright}|p{26ex}<{\raggedright}|}
\hline
\rowcolor{color3}
\texttt{res-m} & \texttt{res-c}\\
\hline
conv2d(16, 3x3)  & conv2d(16, 3x3)\\
$\text{1$*$rb}
\begin{cases}
     \text{conv2d(16, 3x3)}\\
   \text{conv2d(16, 3x3)}
\end{cases}$ 
&  
$\text{3$*$rb}
\begin{cases}
     \text{conv2d(16, 3x3)}\\
   \text{conv2d(16, 3x3)}
\end{cases}$\\
 $\text{1$*$rb}
 \begin{cases}
     \text{conv2d(32, 3x3)}\\
   \text{conv2d(32, 3x3, s=2)}
\end{cases}$ 
& 
$\text{3$*$rb}
 \begin{cases}
     \text{conv2d(32, 3x3)}\\
   \text{conv2d(32, 3x3, s=2)}
\end{cases}$\\
 $\text{1$*$rb}
 \begin{cases}
     \text{conv2d(64, 3x3)}\\
   \text{conv2d(64, 3x3, s=2)}
\end{cases}$ 
&
$\text{3$*$rb}
 \begin{cases}
     \text{conv2d(64, 3x3)}\\
   \text{conv2d(64, 3x3, s=2)}
\end{cases}$\\
softmax(10) & avg-pool(8x8)\\
 & softmax(10) \\
\hline
\end{tabular}
\end{center}
\vspace{-3ex}
\caption{Architectures of naturally trained residual networks for MNIST (\texttt{res-m}) and CIFAR10 (\texttt{res-c}). `$*$' indicates the number of times a layer is repeated. `s' and `rb' represent stride and residual block respectively. Each convolutional layer is followed by a ReLU activation. Batch-norm is used after each convolutional layer in \texttt{res-c}. }
\label{tab:mnist_cifar_res_models}
\end{table}

\subsubsection{Network Architectures}
\textbf{Classification:} 
We study eight models trained on the ImageNet dataset \cite{ILSVRC15}. These models can be grouped into two categories. \textit{(a) Naturally trained:} Five of these models are only trained on benign examples. These include Inceptionv3 (\texttt{Inc-v3}) \cite{szegedy2016rethinking}, Inceptionv4 (\texttt{Inc-v4}),  Inception Resnet v2 (\texttt{IncRes-v2})  \cite{szegedy2017inception} and  Resnet v2-152 (\texttt{Res-152}) \cite{he2016identity} and VGG-19 \cite{Simonyan14verydeep}. 
\textit{(b) Adversarially trained:} The other three, models including \texttt{Adv-v3} \cite{kurakin2016adversarial}, \texttt{Inc-v3\textsubscript{ens3}} and \texttt{IncRes-v2\textsubscript{ens}} \cite{tramer2017ensemble}, are adversarially trained and made publicly available. The specific details about these models can be found in \cite{kurakin2016adversarial, tramer2017ensemble}. Attacks are created for naturally trained models, while tested against all of them. For classification on smaller datasets, we study three models each for MNIST and CIFAR10. Among these models, two  are naturally trained and one is adversarially trained using saddle point optimization \cite{madry2018towards}. Adversarial examples are created for naturally trained models, named \texttt{model-m} and \texttt{model-c} for MNIST and CIFAR10, respectively (see Table \ref{tab:mnist_cifar_conv_models}). These examples are subsequently transferred to adversarially trained Madry's models \cite{madry2018towards} and naturally trained ResNet models, named \texttt{res-m} and \texttt{res-c} for MNIST and CIFAR10 respectively (see Table \ref{tab:mnist_cifar_res_models}).

\begin{table*}[!htp]
\footnotesize
\begin{center}
\begin{tabular}{|c|p{9ex}<{\centering}|p{9ex}<{\centering}|p{9ex}<{\centering}|p{11ex}<{\centering}|p{9ex}<{\centering}|p{9ex}<{\centering}|p{11ex}<{\centering}|c|}
\hline \rowcolor{color3}
Accuracy &\multicolumn{5}{c|}{Naturally Trained}&\multicolumn{3}{c|}{Adv. Trained}\\
\cline{2-9} \rowcolor{color3}
 & \texttt{Inc-v3} & \texttt{Inc-v4} & \texttt{Res-152} & \texttt{IncRes-v2} & \texttt{VGG-19} & \texttt{Adv-v3} & \texttt{Inc-v3\textsubscript{ens3}} & \texttt{IncRes-v2\textsubscript{ens}} \\
\hline\hline
T-1&95.3&\sbest{97.7}&96.1&\best{100.0}&85.5&\sbest{94.3}&90.2&\best{96.9}\\
\hline
T-5&99.8&99.8&\sbest{99.9}&\best{100.0}&96.7&\sbest{99.4}&95.5&\best{99.8}\\
\hline
\end{tabular}
\end{center}
\vspace{-3ex}
\caption{Model accuracies are reported on original data set ImageNet-NIPS containing benign examples only. T-1: top-1 and T-5: top-5 accuracies. Best and second best performances are colorized.}
\label{tab:imagenet_nips_accs}
\end{table*}

\begin{table*}[!htp]
\footnotesize
\begin{center}
\begin{tabular}{|l|p{14ex}<{\raggedright}|p{3.8ex}<{\centering} p{3.8ex}<{\centering}|p{3.8ex}<{\centering} p{3.8ex}<{\centering}|p{3.8ex}<{\centering} p{3.8ex}<{\centering}|p{3.8ex}<{\centering} p{3.8ex}<{\centering}|p{3.8ex}<{\centering} p{3.8ex}<{\centering}|p{3.8ex}<{\centering} p{3.8ex}<{\centering}|p{3.8ex}<{\centering} p{3.8ex}<{\centering}|p{3.8ex}<{\centering} c|}
\hline
\rowcolor{color3}
& &\multicolumn{10}{c|}{Naturally Trained}&\multicolumn{6}{c|}{Adv. Trained}\\
\cline{3-18}\rowcolor{color3}
& Attack & \multicolumn{2}{c|}{Inc-v3}  & \multicolumn{2}{c|}{Inc-v4} & \multicolumn{2}{c|}{Res-152} & \multicolumn{2}{c|}{IncRes-v2} & \multicolumn{2}{c|}{VGG-19} & \multicolumn{2}{c|}{Adv-v3} & \multicolumn{2}{c|}{Inc-v3\textsubscript{ens3}} & \multicolumn{2}{c|}{IncRes-v2\textsubscript{ens}} \\
\cline{3-18}\rowcolor{color3}
& & T-1 & T-5&T-1&T-5&T-1&T-5&T-1&T-5&T-1&T-5&T-1&T-5&T-1&T-5&T-1&T-5\\
\hline\hline

\multirow{5}{*}{\rotatebox[origin=c]{90}{\parbox[c]{1.5cm}{\centering\texttt{Inc-v3}}}} & FGSM \cite{goodfellow2014explaining} & 22.0$^*$ &45.7$^*$&62.5&84.7&64.6&85.8&65.9&85.9&49.9&75.7&69.1&88.1&77.2&90.9& 90.8& 98.3  \\
& R-FGSM \cite{tramer2017ensemble} & 16.7$^*$ &38.0$^*$&65.8&86.0&69.5&89.7&68.8&88.7& 61.4&83.8&76.4&90.9&77.9&91.2&88.8&97.6\\
& I-FGSM \cite{Kurakin2016AdversarialEI}& 0.0$^*$ &1.7$^*$&82.0&97.6&86.5&98.6&90.6&99.1&76.7&95.0&88.5&98.7&84.9&94.4&94.6&99.5\\
& MI-FGSM \cite{Dong_2018_CVPR}&  0.0$^*$ &1.5$^*$&47.1&78.8&47.1&84.5&52.5&81.9&47.3&76.7&71.6&89.8&73.8&90.7&88.3&98.0  \\
& TAP \cite{Zhou_2018_ECCV} & 0.0$^*$ & - & 22.1 & - & 46.9 & - & 24.7 & - & -&- & 52.5&-&60.9&-&68.8&-\\
& DIM \cite{xie2018improving}& 0.2$^*$ &1.3$^*$&27.8&63.1&42.1&75.2&34.6&65.4&40.2&71.4&65.2&87.9&68.3&89.6&86.3&97.5  \\
\hline
\multirow{5}{*}{\rotatebox[origin=c]{90}{\parbox[c]{1.5cm}{\centering\texttt{Res-152}}}} & FGSM \cite{goodfellow2014explaining} &54.1&79.3&61.2&84.2&16.5$^*$&41.0$^*$&62.5&85.6&46.0&72.7&67.3&87.4&74.0&89.4&88.4&97.7\\
& R-FGSM \cite{tramer2017ensemble}&58.5&83.4&64.9&86.6& 12.9$^*$&35.2$^*$&69.1&88.5& 56.1&80.8&74.5&90.6&75.5&90.4& 86.5&96.5 \\
& I-FGSM \cite{Kurakin2016AdversarialEI} &80.0&96.6&84.1&98.4&0.9$^*$&6.2$^*$&92.5&99.1&75.7&94.9&87.4&99.0&85.5&94.8&93.4&99.3 \\
& MI-FGSM \cite{Dong_2018_CVPR}&43.5&76.8&49.9&79.2&0.9$^*$&5.1$^*$&54.8&82.4&46.8&76.0&72.6&90.7&71.1&90.1&86.0&97.5  \\
& TAP \cite{Zhou_2018_ECCV} &48.2&-&55.7&-&7.6$^*$&-&55.2&-&-&-&49.2&-&57.8&-&64.1&-\\
&DIM \cite{xie2018improving} &20.1&51.2&22.0&54.6&0.6$^*$&4.2$^*$&24.6&57.3&33.3&62.6&53.5&82.5&55.2&83.1&74.4&94.1  \\
\hline
\multirow{5}{*}{\rotatebox[origin=c]{90}{\parbox[c]{1.5cm}{\centering\texttt{IncRes-v2}}}}  & FGSM \cite{goodfellow2014explaining} &61.7&83.8&69.6&87.6&68.4&89.6&50.1$^*$&73.9$^*$&52.3&76.5&72.0&89.6&79.0&91.6&90.0&97.7\\
& R-FGSM \cite{tramer2017ensemble}&66.6&87.0&71.8&89.4&73.5&91.5&46.1$^*$&71.3$^*$&62.9&84.1&75.5&91.2&79.3&91.5&87.4&97.3\\
& I-FGSM \cite{Kurakin2016AdversarialEI} &62.8&88.4& 68.3&91.9&77.2&94.8&1.1$^*$&2.6$^*$&71.4&91.7&85.6&97.5&83.8&95.6&89.8&98.4 \\
& MI-FGSM \cite{Dong_2018_CVPR} &36.0&67.5&42.4&73.2&49.3&82.2&1.0$^*$&2.4$^*$&51.3&76.8&70.0&90.1&71.5&92.2&81.8&96.3\\
& TAP \cite{Zhou_2018_ECCV}&25.9&-&33.2&-&53.5&-&4.8$^*$&-&-&-&60.5&-&79.1&-&87.8&- \\
& DIM \cite{xie2018improving} &21.4&49.8&23.5&53.4&32.3&64.3&4.8$^*$&13.7$^*$&39.7&69.2&54.9&81.4&57.5&85.9&73.5&94.4 \\
\hline
\multirow{7}{*}{\rotatebox[origin=c]{90}{\parbox[c]{1.5cm}{\centering\texttt{VGG16}}}} &FGSM \cite{goodfellow2014explaining}&30.1&56.0&34.0&58.0&36.6&65.2&42.2&66.1&9.1&27.9&48.8&72.6&53.5&79.5&72.8&91.1\\
&R-FGSM \cite{tramer2017ensemble}&41.5&67.9&45.1&72.5&49.2&78.4&54.9&77.7&12.9&35.8&63.9&86.2&63.5&85.3&77.1&93.0\\
&I-FGSM \cite{Kurakin2016AdversarialEI}&69.4&93.0&75.3&94.5&79.5&95.7&87.2&97.9&18.3&56.1&82.2&97.5&80.9&93.7&91.5&99.1\\
&MI-FGSM \cite{Dong_2018_CVPR}&16.9&42.0&18.7&40.1&24.9&51.6&26.1&52.5&2.0&14.4&38.8&68.1&42.5&\sbest{72.4}&64.2&\sbest{87.7}\\
& TAP \cite{Zhou_2018_ECCV} & 23.9 & - & 28.1&-&23.9&-&32.3&-&-&-&38.8&-&\sbest{41.9}&-&\sbest{63.8}&-\\
&DIM \cite{xie2018improving}&12.9&35.5&15.2&35.8&20.6&45.7&19.7&43.8&\best{0.6}&\best{8.8}&31.6&59.0&\best{32.1}&\best{61.0}&\best{56.3}&\best{81.0}\\
\cline{2-18}
&FFF \cite{mopuri-bmvc-2017} &61.7&80.7&60.8&78.7&72.8&90.1&76.1&90.1&44.0&68.0&79.6&93.1&83.1&93.1&92.8&98.5  \\
\cline{2-18}
&NRDM&\sbest{5.1}&\best{10.2}&\sbest{6.2}&\sbest{12.4}&\best{15.6}&\best{27.6}&\sbest{13.6}&\sbest{23.0}&\sbest{4.5}&14.2&\best{27.7}&\sbest{46.8}&54.2&75.4&75.3&89.8\\
&NRDM-DIM&\best{4.9}&\sbest{10.5}&\best{5.7}&\sbest{12.0}&\sbest{16.0}&\sbest{28.6}&\best{12.7}&\best{22.6}&5.0&\sbest{14.0}&\sbest{28.7}&\sbest{45.7}&52.9&73.8&74.0&89.5\\
\hline
\end{tabular}
\end{center}
\vspace{-3ex}
\caption{Model accuracies are reported under untargeted $l_\infty$ adversarial attacks on ImageNet-NIPS with perturbation budget $l_\infty \le 16$ for pixel space [0-255]. T-1 and T-2 represent top-1 and top-5 accuracies, respectively. NRDM shows higher or competitive success rates for black-box models than FGSM \cite{goodfellow2014explaining}, I-FGSM \cite{Kurakin2016AdversarialEI}, MI-FGSM \cite{Dong_2018_CVPR}, TAP \cite{Zhou_2018_ECCV}, DIM \cite{xie2018improving} and FFF \cite{mopuri-bmvc-2017}. NRDM-DIM combines input diversity as well as momentum with NRDM.  `$^*$' indicates the white-box attacks. Best and second best black-box attacks are colorized. }
\label{tab:untarget_imagenet}
\end{table*}

\textbf{Object Detection:} To demonstrate cross-task and cross-dataset transferability, we study naturally trained \texttt{RetinaNet} \cite{lin2018focal} performance against adversarial examples found by the NRDM approach (Algorithm \ref{alg:NRD}) on the MS-COCO validation set.

\textbf{Segmentation:} We evaluate the robustness of naturally trained \texttt{SegNet-basic} \cite{Badrinarayanan2017SegNetAD} against adversarial examples generated by the NRDM approach (Algorithm \ref{alg:NRD}) on the CAMVID \cite{brostow2009semantic} test set.

\subsubsection{Attack Parameters}

\begin{figure}[!htp]
  \centering
    \includegraphics[trim= 7mm 2mm 14mm 14mm, clip, width=.85\linewidth]
    {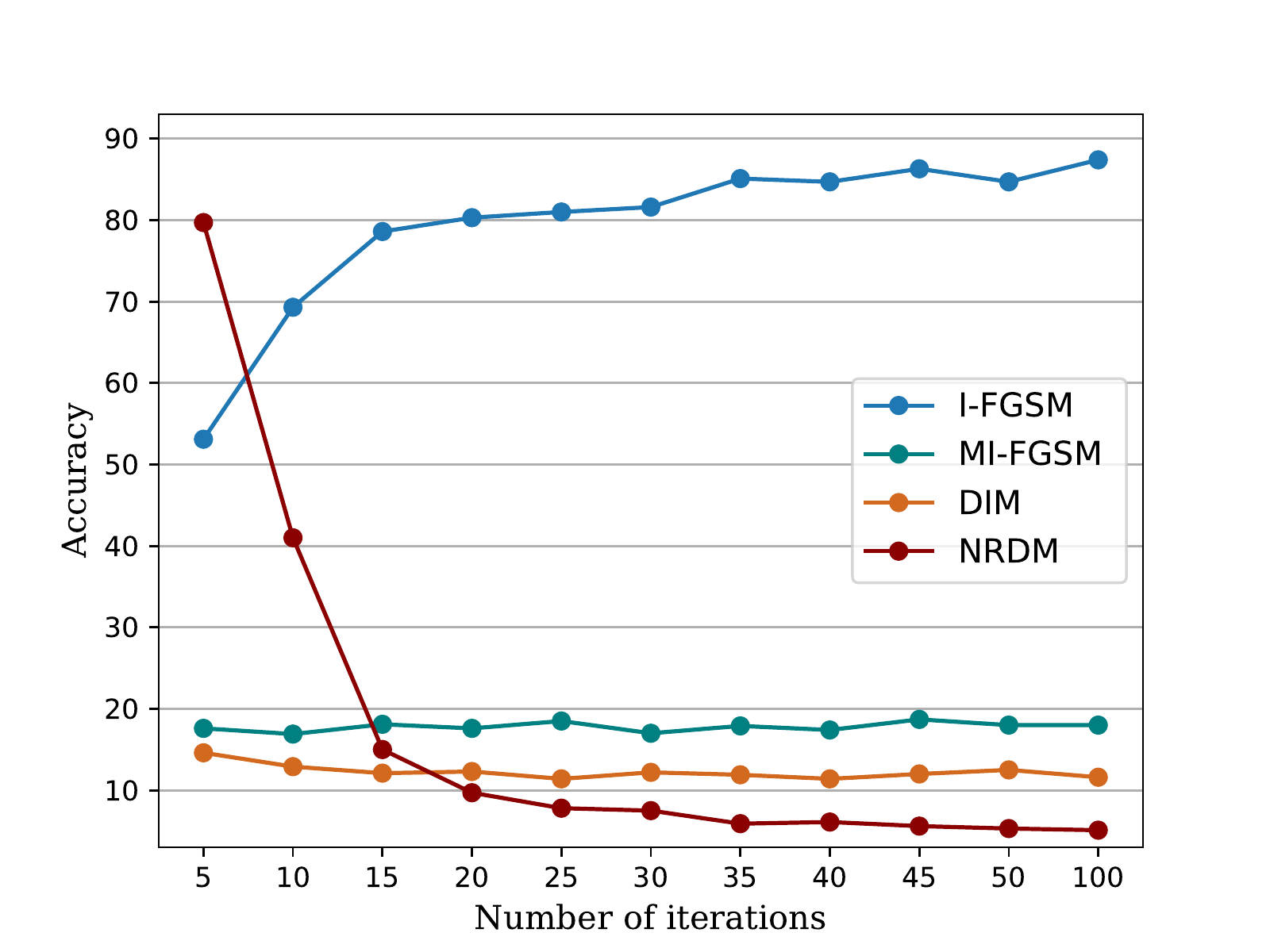}
    \caption{Accuracy of Inc-v3 for adversarial examples generated on VGG-16 by I-FGSM and MI-FGSM, DIM and NRDM. NRDM's strength increases with number of iterations, in contrast to MI-FGSM and DIM.}
    \label{fig:NRD_iterations}
    \vspace{-2ex}
\end{figure}

FGSM is a single-step attack. Its step size is set to $16$. In the case of R-FGSM, we take a step of size $\alpha{=}16/3$ in a random direction and then a gradient step of size $16{-}\alpha$ to maximize model loss. The attack methods, I-FGSM, MI-FGSM and DIM, are run for ten iterations. The step size for these attacks is set to 1.6, as per standard practice. The momentum decay factor for MI-FGSM is set to one. This means that attack accumulates all the previous gradient information to perform the current update and is shown to have the best success rate \cite{Dong_2018_CVPR}. For DIM, the transformation probability is set to 0.7. In the case of FFF \cite{mopuri-bmvc-2017}, we train the adversarial noise for 10K iterations to maximize the response at the activation layers of VGG-16 \cite{Simonyan14verydeep}. For the NRDM (Algorithm \ref{alg:NRD}),  we used the VGG-16 \cite{Simonyan14verydeep} conv3-3 feature map as the representation loss. Since NRDM maximizes loss w.r.t a benign example, it does not suffer from the over-fitting problem. We run NRDM for the maximum number of 100 iterations. The transferability of different attacks is compared against the number of iterations in Fig.~\ref{fig:NRD_iterations}. MI-FGSM and DIM quickly reach to their full potential within ten iterations. The strength of I-FGSM strength decreases, while NRDM strength increases, with the number of attack iterations.

\begin{table}[!htp]
\footnotesize
\begin{center}
\begin{tabular}{|c|p{10ex}<{\raggedright}|p{9ex}<{\centering}|p{9ex}<{\centering}|c|}
\hline \rowcolor{color3}
Datasets $\downarrow$ & Attack $\downarrow$ &\multicolumn{2}{c|}{Naturally Trained}&\multicolumn{1}{c|}{Adv. Trained}\\
\cline{3-5}\rowcolor{color3}
&  & \texttt{model-m} & \texttt{res-m} & \texttt{Madry-M}  \\
\hline\hline

\multirow{4}{*}{MNIST} & FGSM & 42.28$^*$ & 53.15 & \best{95.96} \\
& I-FGSM& \sbest{40.66}$^*$ & 51.04 &  \sbest{96.64}  \\
& MI-FGSM & \sbest{40.66}$^*$ & \sbest{48.19} & \best{95.96} \\
& NRDM & \best{4.39}$^*$ & \best{23.54}$^*$ & 97.56 \\
\hline\rowcolor{color3}
\multicolumn{2}{|c|}{} & \texttt{model-c}& \texttt{res-c} & \texttt{Madry-C} \\
\hline
\multirow{4}{*}{CIFAR10} & FGSM & \sbest{5.47}$^*$ & 24.19 & \best{85.54} \\
& I-FGSM & \best{2.52}$^*$ & 36.81 & 87.00 \\
& MI-FGSM& \best{2.52}$^*$ & \best{16.56} &  \sbest{85.71} \\
& NRDM & 11.92$^*$ & \sbest{23.98} & 86.99\\
\hline
\end{tabular}
\end{center}
\vspace{-3ex}
\caption{Model accuracies under untargeted $l_\infty$ adversarial attacks on MNIST and CIFAR10 with perturbation budget $l_\infty \le 76.5$  and $l_\infty \le 8$, respectively, for pixel space [0-255], as per standard practice \cite{madry2018towards}. NRDM shows higher or competitive success rates for black-box models compared to FGSM, I-FGSM and MI-FGSM. `$^*$' indicates the white-box attacks. Best and second best attacks are colorized.}
\label{tab:untarget_mnist/cifar10}
\end{table}

\begin{table}[!htp]
\footnotesize
\begin{center}
\begin{tabular}{|l|p{9ex}<{\raggedright}|p{9ex}<{\centering}|p{9ex}<{\centering}|c|}
\hline \rowcolor{color3}
Dataset $\downarrow$ & Metric $\downarrow$ &\multicolumn{2}{c|}{Naturally Trained}&\multicolumn{1}{c|}{Adv. Trained}\\
\cline{3-5}
\rowcolor{color3}
&  & \texttt{model-m} & \texttt{res-m} & \texttt{Madry-M}  \\
\hline\hline

MNIST & Accuracy & \best{99.30} & \sbest{98.88} & 98.40 \\
\hline \rowcolor{color3}
\multicolumn{2}{|c|}{} & \texttt{model-c}& \texttt{res-c} & \texttt{Madry-C} \\
\hline 
CIFAR10 & Accuracy & \sbest{85.44} & 80.56 & \best{87.62} \\
\hline
\end{tabular}
\end{center}
\vspace{-3ex}
\caption{Model accuracies on original test datasets for MNIST and CIFAR10 containing benign examples only. Best and second best performances are colorized.}
\label{tab:mnist_cifar10_accs}
\end{table}

\subsection{Input Transformations}
\label{sec: transformations}
Different input transformations have been proposed to mitigate the adversarial effect but they can be easily broken in a \emph{white-box} scenario. This is because an attacker can be adaptive and incorporate transformations into the adversary generation process. Even non-differentiable transformations can be by-passed by approximating them with an identity function \cite{athalye2018obfuscated}. However in a \emph{black-box} scenario, the attacker does not have any knowledge of the transformation function along with the network architecture and its parameters. We test the strength of our adversarial attack against well studied transformations, including JPEG, total variation minimization (TVM) and median filtering. We report our experimental results using the above-mentioned network architectures and input transformations in the following section.

\begin{table}[!htp]
\footnotesize
\begin{center}
\begin{tabular}{|l|p{10ex}<{\centering}|p{9ex}<{\centering}|c|}
\hline \rowcolor{color3}
Method & No Attack &\multicolumn{2}{c|}{NRDM}\\
\cline{3-4} \rowcolor{color3}
&  & $l_\infty \le 8$ & $l_\infty \le 16$   \\
\hline\hline
No Defense & \best{79.70} & 52.48 & 32.59  \\
\hline
JPEG (quality=75) & \sbest{77.25}&51.76 & 32.44\\
JPEG (quality=50) & 75.27& 52.45& 33.16\\
JPEG (quality=20) & 68.82 & \sbest{53.08}& \best{35.54} \\
\hline
TVM (weights=30) & 73.70 & 55.54 & 34.21 \\ 
TVM (weights=10) & 70.38 & \best{59.52} & \sbest{34.57}\\ 
\hline
MF (window=3) & 75.65 & 49.18 & 30.52\\ 
\hline
\end{tabular}
\end{center}
\vspace{-4ex}
\caption{Segnet-Basic accuracies on CAMVID test set with and without input transformations against NRDM. Best and second best performances are colorized.}
\label{tab:segnet}
\end{table}

\begin{table}[!htp]
\footnotesize
\begin{center}
\begin{tabular}{|l|p{10ex}<{\centering}|p{9ex}<{\centering}|c|}
\hline \rowcolor{color3}
Method & No Attack &\multicolumn{2}{c|}{NRDM}\\
\cline{3-4}  \rowcolor{color3}
&  & $l_\infty \le 8$ & $l_\infty \le 16$   \\
\hline\hline
No Defense & \best{53.78}  & 22.75 & 5.16  \\
\hline
\hline
JPEG (quality=75) & \sbest{49.57} & 20.73 & 4.7\\
JPEG (quality=50) & 46.36 & 19.89 & 4.33 \\
JPEG (quality=20) & 40.04 & 19.13 & 4.58\\
\hline
TVM (weights=30) & 47.06 & \sbest{27.63} & \sbest{6.36} \\ 
TVM (weights=10) & 42.79 & \best{32.21} & \best{9.56}\\ 
\hline
MF (window=3) & 43.48 & 19.59 & 5.05\\ 
\hline
\end{tabular}
\end{center}
\vspace{-4ex}
\caption{mAP (with IoU = 0.5) of RetinaNet is reported on the MS-COCO validation set with and without input transformations against NRDM. Best and second best performances are colorized.}
\label{tab:object_detection}
\end{table}

\begin{table*}[tp]
\footnotesize
\begin{center}
\begin{tabular}{|l|p{5ex}<{\centering} p{5ex}<{\centering}|p{4ex}<{\centering} p{4ex}<{\centering}|p{4ex}<{\centering} p{4ex}<{\centering}|p{4ex}<{\centering} p{4ex}<{\centering}|p{5ex}<{\centering} p{5ex}<{\centering}|p{4ex}<{\centering} p{4ex}<{\centering}|p{4ex}<{\centering} c|}
\hline \rowcolor{color3}
& \multicolumn{2}{c|}{No Attack} & \multicolumn{2}{c|}{FGSM \cite{goodfellow2014explaining}} & \multicolumn{2}{c|}{R-FGSM \cite{tramer2017ensemble}} & \multicolumn{2}{c|}{I-FGSM \cite{Kurakin2016AdversarialEI}} & \multicolumn{2}{c|}{MI-FGSM \cite{Dong_2018_CVPR}} & \multicolumn{2}{c|}{DIM \cite{xie2018improving}} & \multicolumn{2}{c|}{NRDM}\\ 
\cline{2-15} \rowcolor{color3}
&T-1&T-5&T-1&T-5&T-1&T-5&T-1&T-5&T-1&T-5&T-1&T-5&T-1&T-5\\
\hline
No Defense &\best{95.3}&\best{99.8}&30.1&56.0&41.5&\sbest{67.9}&69.4&93.0&16.9&\sbest{42.0}&12.9&35.5&5.1&10.2  \\
\hline
\hline
JPEG (quality=75) & \sbest{93.9}&\sbest{99.5}&30.4&55.4&\sbest{41.8}&67.0&69.7&92.3&\sbest{18.4}&\sbest{42.0}&13.5&33.3&5.4&12.6\\
JPEG (quality=50) & 91.3&99.3&\sbest{31.0}&55.4&40.5&65.3&68.7&91.8&18.1&42.1&13.1&34.4&6.5&12.8\\
JPEG (quality=20) &86.0&97.6&29.9&53.9&38.0&64.6&69.8&90.9&\sbest{18.4}&42.1&14.1&34.2&8.4&\sbest{18.7} \\
\hline
TVM (weights=30) &93.1&99.4&30.6&\sbest{56.2}&41.7&67.7&\sbest{73.7}&\best{94.5}&17.2&42.1&14.9&33.5&9.8&18.5\\ 
TVM (weights=10) & 88.8&97.6&\best{32.1}&\best{57.3}&\best{43.6}&\best{69.4}&\best{73.9}&\sbest{93.4}&\best{19.8}&\best{45.7}&\sbest{15.8}&\sbest{37.1}&\best{24.0}&\best{40.5}\\ 
\hline
MF (window=3) &93.2&99.1&24.3&45.5&36.1&62.3&62.8&89.9&16.2&36.8&\best{18.8}&\best{42.1}&\sbest{9.9}&17.9 \\ 
\hline
\end{tabular}
\end{center}
\vspace{-4ex}
\caption{Inc-v3 accuracy is reported with and without input transformations. Adversarial examples are generated for VGG-16 in white-box setting by FGSM, R-FGSM, I-FGSM, MI-FGSM, DIM and NRDM under perturbation budget $l_\infty \le 16$ and then transferred to Inc-v3. T-1 and T-2 represent top-1 and top-5 accuracies, respectively. Best and second best performances are colorized.}
\label{tab:transforms_classify}
\end{table*}

\begin{figure*}[!htp]
\centering
  \begin{minipage}{0.24\textwidth}
  	\centering
     \vspace{1em}
    \includegraphics[width=\linewidth, height=3cm]{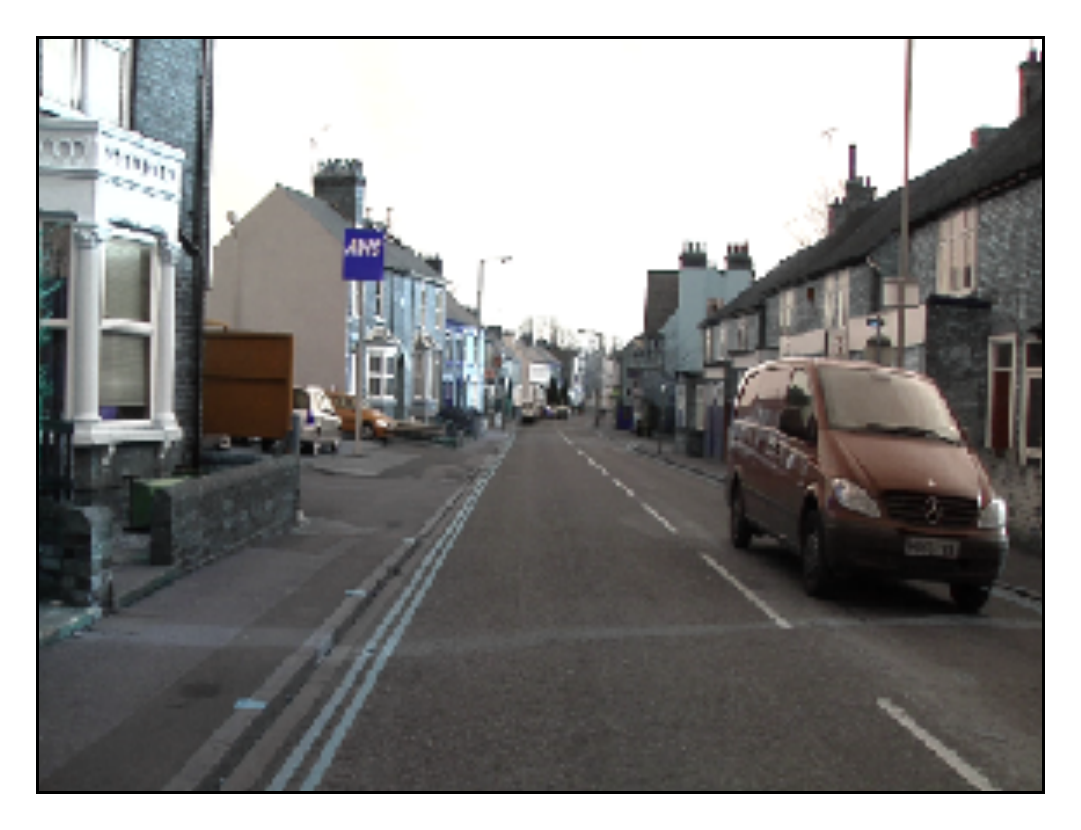}\
     (a) \small Original
     \vspace{1em}
  \end{minipage}
   \begin{minipage}{0.24\textwidth}
   	\centering
    \vspace{1em}
    \includegraphics[width=\linewidth, height=3cm]{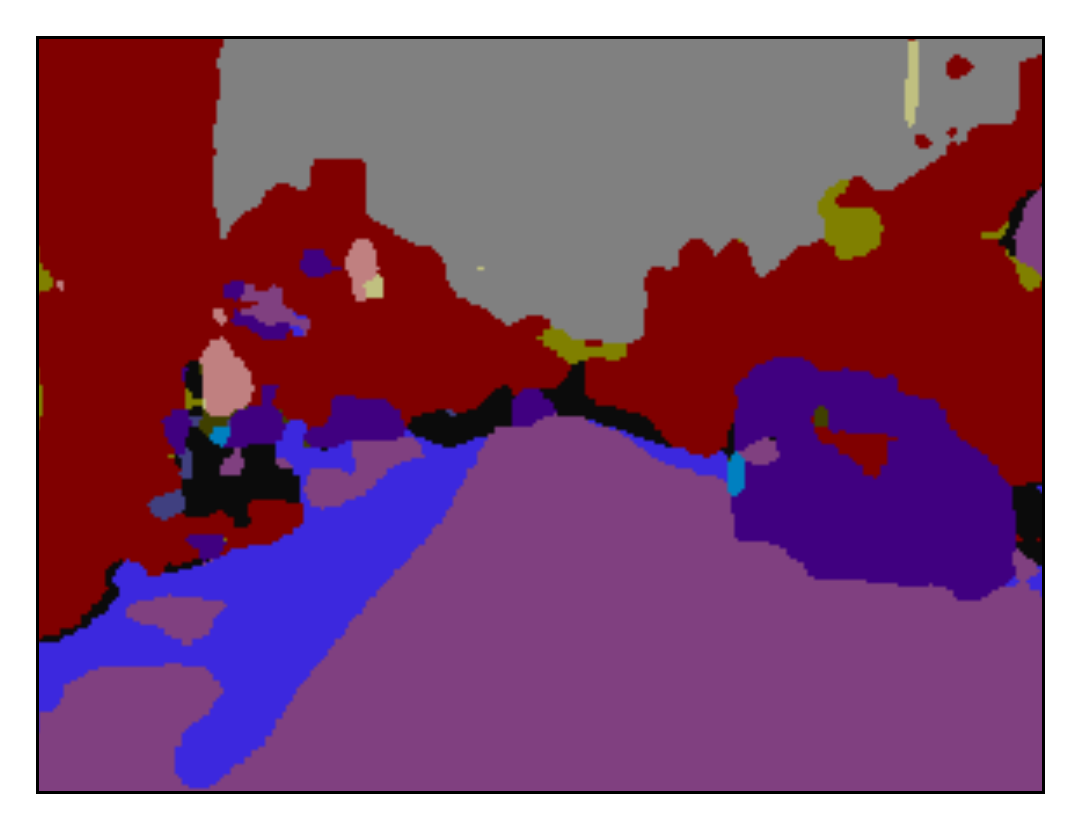}\
      (b) \small Prediction for Original
      \vspace{1em}
  \end{minipage}
  \begin{minipage}{0.24\textwidth}
  	\centering
     $l_\infty \le 16$ 
    \includegraphics[width=\linewidth, height=3cm]{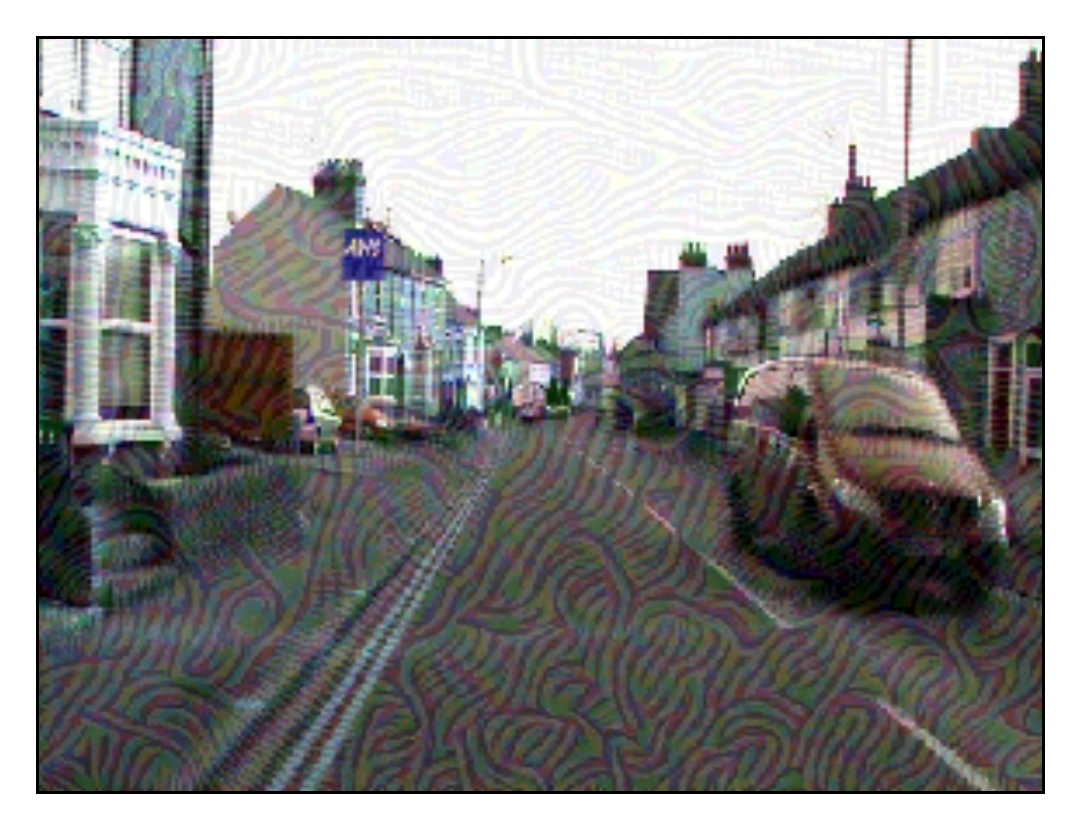}\
     (c) \small Adversarial 
     \vspace{1em}
  \end{minipage}
   \begin{minipage}{0.24\textwidth}
   	\centering
    \vspace{1em}
    \includegraphics[width=\linewidth, height=3cm]{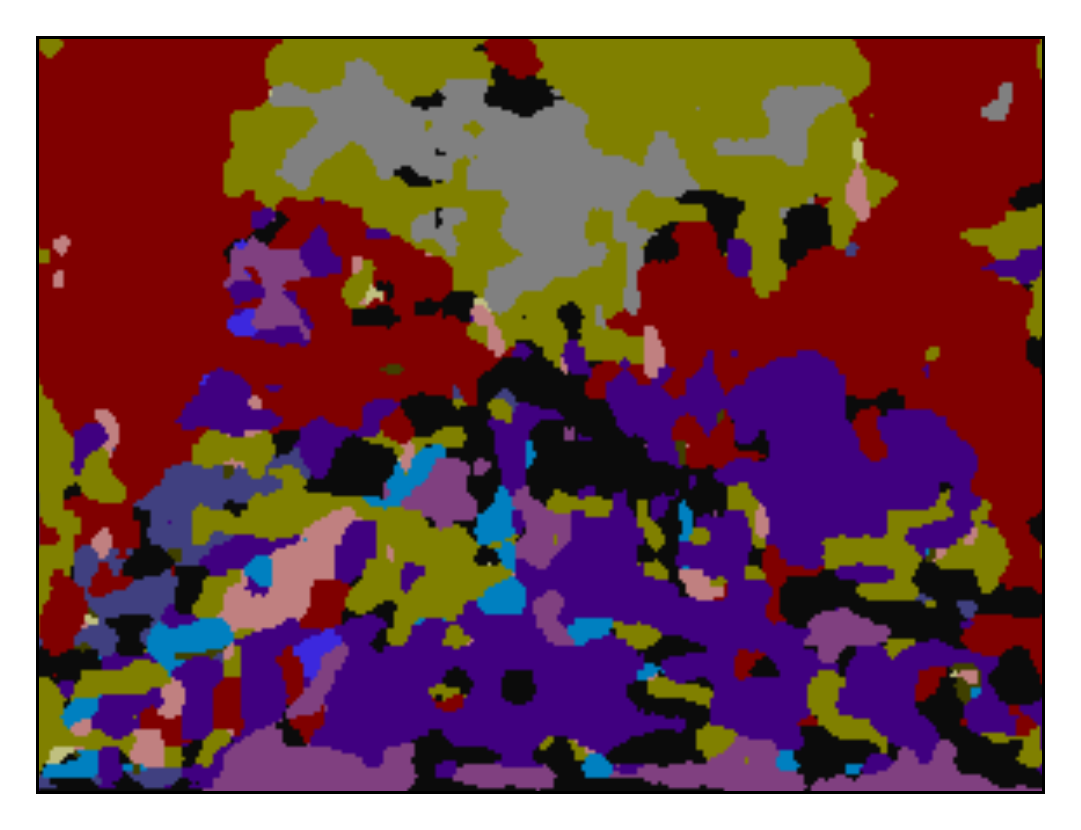}\
      (d) \small Prediction for Adversarial
      \vspace{1em}
  \end{minipage}
  \vspace{-0.4cm}
  \caption{Segnet-Basic output is shown for different images. (a) is the original image, while (b) shows predictions for the original image. (c) is the adversary found by NRDM algorithm \ref{alg:NRD}, while (d) shows predictions for the adversarial image. The perturbation budget is written on the top of adversarial image.}
\label{fig:seg_images}
\end{figure*}

\begin{figure*}[!htp]
\centering
  \begin{minipage}{0.24\textwidth}
  	\centering
    \vspace{1em}
    \includegraphics[width=\linewidth, height=3cm]{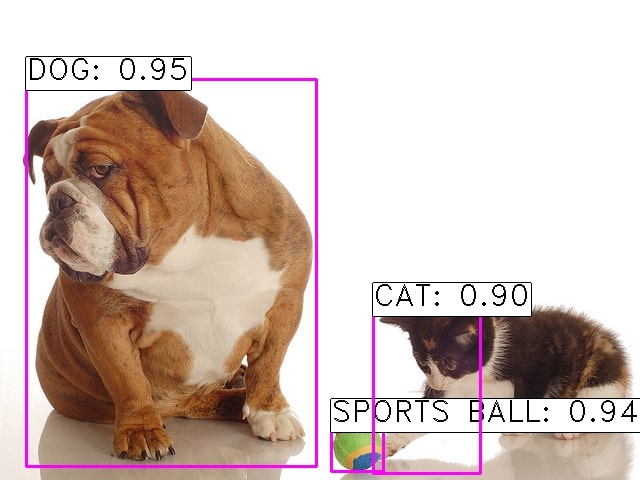}\
     (a) \small  Original
     \vspace{1em}
  \end{minipage}
   \begin{minipage}{0.24\textwidth}
   	\centering
    $l_\infty \le 8$
    \includegraphics[width=\linewidth, height=3cm]{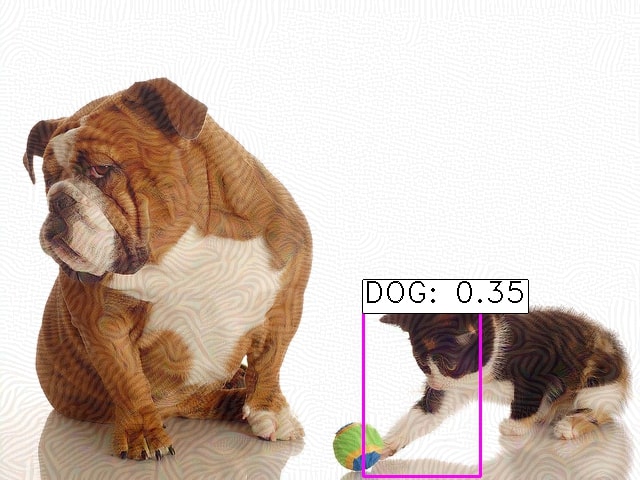}\
      (b) \small  Adversarial
      \vspace{1em}
  \end{minipage}
  \begin{minipage}{0.24\textwidth}
  	\centering
    \vspace{1em}
    \includegraphics[width=\linewidth, height=3cm]{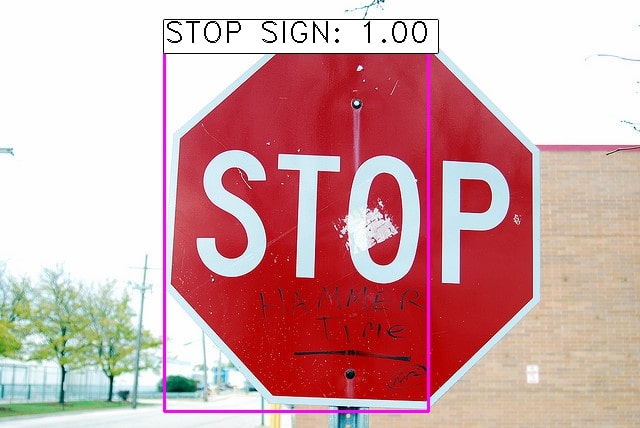}\
     (c) \small  Original
     \vspace{1em}
  \end{minipage}
   \begin{minipage}{0.24\textwidth}
   	\centering
    $l_\infty \le 16$
    \includegraphics[width=\linewidth, height=3cm]{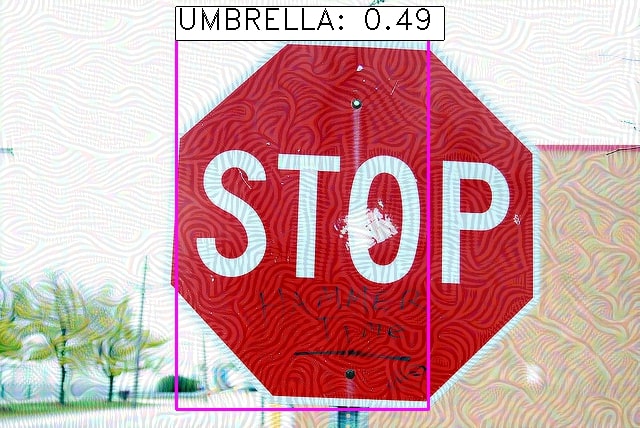}\
      (d) \small  Adversarial
      \vspace{1em}
  \end{minipage}
  \vspace{-0.4cm}
  \caption{RetinaNet detection results are shown for different images. (a) and (c) show detection for the original images, while (b) and (d) show detection for adversaries found using NRDM algorithm \ref{alg:NRD}. The perturbation budget is written on the top of each adversarial image.}
\label{fig:detect_images}
\end{figure*}

\section{Results}
\label{sec: results}
\textbf{Classification:} We report the performance of our attack against a number of CNN architectures on the ImageNet-NIPS dataset in Table \ref{tab:untarget_imagenet}. The following insights can be drawn from our results. \textbf{(1)} In comparison to other state-of-the-art attacks, our approach consistently demonstrates a much higher transferability rate for naturally trained images.  Specifically, NRDM attack have much higher transferability on naturally trained models, bringing down top-1 accuracy of \texttt{IncRes-v2} \cite{szegedy2017inception} from 100.0\% (see Table \ref{tab:imagenet_nips_accs}) to 12.7\% (see Table \ref{tab:untarget_imagenet}). \textbf{(2)} In comparison, MI-FGSM \cite{Dong_2018_CVPR} and DIM \cite{xie2018improving} perform slightly better on adversarially trained ensemble models \cite{tramer2017ensemble}, with NRDM showing competitive success rate. This is because the MI-FGSM and DIM methods use decision boundary information while, NRDM is agnostic to decision-level information about the classifier. \textbf{(3)} We also test with adversarial examples found using different network architectures (i.e., \texttt{Inc-v3}, \texttt{Res-152}, \texttt{IncRes-v2}, \texttt{VGG16}). Overall, we conclude that the adversarial examples found in VGG-16 \cite{Simonyan14verydeep} space have very high transferability. Figure \ref{fig:L_inf_images} shows a visual comparison of adversaries found by different attack algorithms. On small datasets (MNIST and CIFAR10), similar to other attacks, the NRDM becomes ineffective against adversarially trained \texttt{Madry}  models \cite{madry2018towards} (see Tables \ref{tab:mnist_cifar10_accs} and \ref{tab:untarget_mnist/cifar10}) in black-box settings. This shows that finding better methods for adversarial training is a way forward to defend against these attacks. Input transformations can somewhat help to mitigate the adversarial effect in black-box settings (see Table \ref{tab:transforms_classify}). TVM is the most effective against all the attacks, while median filtering perform better against DIM \cite{xie2018improving}. JPEG is the least effective against untargeted adversarial attacks.

\textbf{Segmentation:}
 The NRDM attack created on CAMVID \cite{brostow2009semantic} in VGG-16 feature space is able to bring down the per pixel accuracy of Segnet-Basic by 47.11\% within $l_\infty \le 16$ (see Table \ref{tab:segnet} and Fig \ref{fig:seg_images}). JPEG and TVM transformations are slightly effective but only at the cost of accuracy on benign examples.

\textbf{Object Detection:} RetinaNet \cite{lin2018focal} collapses in the presence of adversaries found by NRDM on the MS-COCO validation set using the VGG-16 \cite{Simonyan14verydeep} feature space. Its mean average precision (mAP) with 0.5 intersection over union (IOU) drops from 53.78\% to 5.16\% under perturbation budget $l_\infty \le 16$ (see Table \ref{tab:object_detection} and Fig \ref{fig:detect_images}). TVM is relatively more effective compared to other transforms against the NRDM attack.

\section{Conclusion}
\label{sec:discussion}
We propose a novel attack algorithm to demonstrate how to benefit from generic internal neural representations of pretrained models (e.g., VGG-16) on ImageNet dataset to exhibit cross-architecture, cross-dataset and cross-task transferability. 
The fact that adversarial examples created just by maximizing the perceptual distortion can fool multiple vision systems is both intriguing and puzzling. As compared to Resnet/Inception, VGG networks generally perform lower, showing that learned features are not optimal for the classification task. Despite this, the adversaries found in VGG feature space are the most transferable. This poses a serious challenge, as an attacker can simply train a local copy of a VGG style network and use its adversaries to penetrate any underlying vision system. The Generalizability of our attack algorithm makes it suitable to make any system robust against adversaries via adversarial training that benefits from generic off-the-shelf feature representations of  pre-trained classification models.

{\small
\bibliographystyle{ieee}
\bibliography{egbib}

\begin{thebibliography}{10}\itemsep=-1pt

\bibitem{athalye2018obfuscated}
A.~Athalye, N.~Carlini, and D.~A. Wagner.
\newblock Obfuscated gradients give a false sense of security: Circumventing
  defenses to adversarial examples.
\newblock In {\em International Conference on Machine Learning (ICML)}, 2018.

\bibitem{athalye2017synthesizing}
A.~Athalye, L.~Engstrom, A.~Ilyas, and K.~Kwok.
\newblock Synthesizing robust adversarial examples.
\newblock In {\em International Conference on Machine Learning (ICML)}, 2017.

\bibitem{Badrinarayanan2017SegNetAD}
V.~Badrinarayanan, A.~Kendall, and R.~Cipolla.
\newblock Segnet: A deep convolutional encoder-decoder architecture for image
  segmentation.
\newblock {\em IEEE Transactions on Pattern Analysis and Machine Intelligence},
  39:2481--2495, 2017.

\bibitem{brostow2009semantic}
G.~J. Brostow, J.~Fauqueur, and R.~Cipolla.
\newblock Semantic object classes in video: A high-definition ground truth
  database.
\newblock {\em Pattern Recognition Letters}, 30(2):88--97, 2009.

\bibitem{carlini2017towards}
N.~Carlini and D.~Wagner.
\newblock Towards evaluating the robustness of neural networks.
\newblock In {\em 2017 IEEE Symposium on Security and Privacy (SP)}, pages
  39--57. IEEE, 2017.

\bibitem{Dong_2018_CVPR}
Y.~Dong, F.~Liao, T.~Pang, H.~Su, J.~Zhu, X.~Hu, and J.~Li.
\newblock Boosting adversarial attacks with momentum.
\newblock In {\em The IEEE Conference on Computer Vision and Pattern
  Recognition (CVPR)}, June 2018.

\bibitem{goodfellow2014explaining}
I.~Goodfellow, J.~Shlens, and C.~Szegedy.
\newblock Explaining and harnessing adversarial examples.
\newblock In {\em International Conference on Learning Representations (ICRL)},
  2015.

\bibitem{Kurakin2016AdversarialEI}
I.~J. Goodfellow, J.~Shlens, and C.~Szegedy.
\newblock Adversarial examples in the physical world.
\newblock In {\em International Conference on Learning Representations (ICRL)},
  2017.

\bibitem{he2016identity}
K.~He, X.~Zhang, S.~Ren, and J.~Sun.
\newblock Identity mappings in deep residual networks.
\newblock In {\em European conference on computer vision}, pages 630--645.
  Springer, 2016.

\bibitem{kornblith2018better}
S.~Kornblith, J.~Shlens, and Q.~V. Le.
\newblock Do better imagenet models transfer better?
\newblock {\em arXiv preprint arXiv:1805.08974}, 2018.

\bibitem{Kupyn_2018_CVPR}
O.~Kupyn, V.~Budzan, M.~Mykhailych, D.~Mishkin, and J.~Matas.
\newblock Deblurgan: Blind motion deblurring using conditional adversarial
  networks.
\newblock In {\em The IEEE Conference on Computer Vision and Pattern
  Recognition (CVPR)}, June 2018.

\bibitem{kurakin2016adversarial}
A.~Kurakin, I.~Goodfellow, and S.~Bengio.
\newblock Adversarial machine learning at scale.
\newblock {\em arXiv preprint arXiv:1611.01236}, 2016.

\bibitem{lin2018focal}
T.-Y. Lin, P.~Goyal, R.~Girshick, K.~He, and P.~Doll{\'a}r.
\newblock Focal loss for dense object detection.
\newblock {\em IEEE transactions on pattern analysis and machine intelligence},
  2018.

\bibitem{lin2014microsoft}
T.-Y. Lin, M.~Maire, S.~Belongie, J.~Hays, P.~Perona, D.~Ramanan,
  P.~Doll{\'a}r, and C.~L. Zitnick.
\newblock Microsoft coco: Common objects in context.
\newblock In {\em European conference on computer vision}, pages 740--755.
  Springer, 2014.

\bibitem{long2015fully}
J.~Long, E.~Shelhamer, and T.~Darrell.
\newblock Fully convolutional networks for semantic segmentation.
\newblock In {\em Proceedings of the IEEE conference on computer vision and
  pattern recognition}, pages 3431--3440, 2015.

\bibitem{madry2018towards}
A.~Madry, A.~Makelov, L.~Schmidt, D.~Tsipras, and A.~Vladu.
\newblock Towards deep learning models resistant to adversarial attacks.
\newblock In {\em International Conference on Learning Representations}, 2018.

\bibitem{mopuri-bmvc-2017}
K.~R. Mopuri, U.~Garg, and R.~V. Babu.
\newblock Fast feature fool: A data independent approach to universal
  adversarial perturbations.
\newblock In {\em Proceedings of the British Machine Vision Conference
  ({BMVC})}, 2017.

\bibitem{ILSVRC15}
O.~Russakovsky, J.~Deng, H.~Su, J.~Krause, S.~Satheesh, S.~Ma, Z.~Huang,
  A.~Karpathy, A.~Khosla, M.~Bernstein, A.~C. Berg, and L.~Fei-Fei.
\newblock {ImageNet Large Scale Visual Recognition Challenge}.
\newblock {\em International Journal of Computer Vision (IJCV)},
  115(3):211--252, 2015.

\bibitem{sharif2014cnn}
A.~Sharif~Razavian, H.~Azizpour, J.~Sullivan, and S.~Carlsson.
\newblock Cnn features off-the-shelf: an astounding baseline for recognition.
\newblock In {\em Proceedings of the IEEE conference on computer vision and
  pattern recognition workshops}, pages 806--813, 2014.

\bibitem{Simonyan14verydeep}
K.~Simonyan and A.~Zisserman.
\newblock Very deep convolutional networks for large-scale image recognition,
  2014.

\bibitem{Su2018IsRT}
D.~Su, H.~Zhang, H.~Chen, J.~Yi, P.-Y. Chen, and Y.~Gao.
\newblock Is robustness the cost of accuracy? - a comprehensive study on the
  robustness of 18 deep image classification models.
\newblock {\em CoRR}, abs/1808.01688, 2018.

\bibitem{szegedy2017inception}
C.~Szegedy, S.~Ioffe, V.~Vanhoucke, and A.~A. Alemi.
\newblock Inception-v4, inception-resnet and the impact of residual connections
  on learning.
\newblock In {\em AAAI}, volume~4, page~12, 2017.

\bibitem{szegedy2016rethinking}
C.~Szegedy, V.~Vanhoucke, S.~Ioffe, J.~Shlens, and Z.~Wojna.
\newblock Rethinking the inception architecture for computer vision.
\newblock In {\em Proceedings of the IEEE Conference on Computer Vision and
  Pattern Recognition}, pages 2818--2826, 2016.

\bibitem{szegedy2013intriguing}
C.~Szegedy, W.~Zaremba, I.~Sutskever, J.~Bruna, D.~Erhan, I.~Goodfellow, and
  R.~Fergus.
\newblock Intriguing properties of neural networks.
\newblock In {\em International Conference on Learning Representations (ICRL)},
  2014.

\bibitem{tramer2017ensemble}
F.~Tram{\`e}r, A.~Kurakin, N.~Papernot, D.~Boneh, and P.~McDaniel.
\newblock Ensemble adversarial training: Attacks and defenses.
\newblock In {\em International Conference on Learning Representations (ICRL)},
  2018.

\bibitem{xie2018improving}
C.~Xie, Z.~Zhang, J.~Wang, Y.~Zhou, Z.~Ren, and A.~Yuille.
\newblock Improving transferability of adversarial examples with input
  diversity.
\newblock {\em arXiv preprint arXiv:1803.06978}, 2018.

\bibitem{Zhou_2018_ECCV}
W.~Zhou, X.~Hou, Y.~Chen, M.~Tang, X.~Huang, X.~Gan, and Y.~Yang.
\newblock Transferable adversarial perturbations.
\newblock In {\em The European Conference on Computer Vision (ECCV)}, September
  2018.

\end{thebibliography}
}

\end{document}